# Inferring Heterogeneous Treatment Effects of Crashes on Highway Traffic: A Doubly Robust Causal Machine Learning Approach

Shuang Li[1], Ziyuan Pu[1,2,3*], Zhiyong Cui[4], Seunghyeon Lee[5], Xiucheng Guo[1], Dong Ngoduy[6]

Abstract: Highway traffic crashes exert a considerable impact on both transportation systems and the economy. In this context, accurate and dependable emergency responses are crucial for effective traffic management. However, the influence of crashes on traffic status varies across diverse factors and may be biased due to selection bias. Therefore, there arises a necessity to accurately estimate the heterogeneous causal effects of crashes, thereby providing essential insights to facilitate individual-level emergency decision-making. This paper proposes a novel causal machine learning framework to estimate the causal effect of different types of crashes on highway speed. The Neyman-Rubin Causal Model (RCM) is employed to formulate this problem from a causal perspective. The Conditional Shapley Value Index (CSVI) is proposed based on causal graph theory to filter adverse variables, and the Structural Causal Model (SCM) is then adopted to define the statistical estimand for causal effects. The treatment effects are estimated by Doubly Robust Learning (DRL) methods, which combine doubly robust causal inference with classification and regression machine learning models. Experimental results from 4815 crashes on Highway Interstate 5 in Washington State reveal the heterogeneous treatment effects of crashes at varying distances and durations. The rear-end crashes cause more severe congestion and longer durations than other types of crashes, and the sideswipe crashes have the longest delayed impact. Additionally, the findings show that rear-end crashes affect traffic greater at night, while crash to objects has the most significant influence during peak hours. Statistical hypothesis tests, error metrics based on matched "counterfactual outcomes", and sensitive analyses are employed for assessment, and the results validate the accuracy and effectiveness of our method.

[1] School of Transportation, Southeast University, China
[2] Key Laboratory of Transport Industry of ComprehensiveTransportation Theory (Nanjing Modern Multimodal TransportationLaboratory), Ministry of Transport, PRC
[3] School of Engineering, Monash University, Malaysia
[4] School of Transportation Science and Engineering, Beihang University, China
[5] Department of Transportation Engineering, University of Seoul, Korea
[6] Department of Civil Engineering, Monash University, Australia

# 1. Introduction

Highway traffic crashes significantly impact highway traffic efficiency, resulting in economic and energy losses. In the United States, the National Highway Traffic Safety Administration estimated that the total cost of car crashes was $340 billion in 2019(Blincore et al., 2023). Furthermore, highway congestion cost users a total of $45.84 billion in 2019, with more than 18% of this cost related to incidents. To mitigate the adverse effects of congestion resulting from crashes, many highway safety-oriented investigations have primarily concentrated on calculating average effects for emergency planning purposes(Chung and Recker, 2013; Li et al., 2013; Mannering et al., 2016; Chung, 2017; Ren and Xu, 2024). Nevertheless, considering that the effects of crashes on traffic conditions differ based on factors like pre-crash traffic conditions, location, and time, comprehending heterogeneous effects can significantly enhance the precision of decision-making. Moreover, guaranteeing the credibility of emergency measures necessitates a profound understanding of the causal effects of crashes on traffic conditions, providing crucial support for policymakers.

While conventional regression methods are commonly utilized to identify the association between traffic crashes and influential factors (Xie et al., 2012; Yu and Abdel-Aty, 2014; Dabbour et al., 2020; Wen et al., 2021; Ding and Sze, 2022), these models overlook the selection bias from the observational data (Mannering et al., 2020). Therefore, incorrect interpretations may be obtained due to the selection bias (Pearl, 2014). For instance, rear-end crashes are more likely to occur during peak hours due to more stop-and-go activities. During peak-hours, traffic volume is much higher than non-peak hours which results in lower traffic speed. Thus, the rear-end collisions usually may not trigger an additional reduction of traffic speed. In this case, peak-hours is a confounder (i.e., the variables that both influence treatment and outcome) between crashes and traffic speed, which leads to underestimating the impact of rear-end crashes on traffic speed reduction. Since numerous confounders may influence the traffic speed and the crash, it is crucial to apply causal inference models to mitigate confounding bias and estimate the causal effects of traffic crashes(Karwa et al., 2011).

Causal inference enables the identification of causal relationships between outcomes and treatments, allowing for the analysis of how the outcomes respond when the causal factors are altered (Morgan and Winship, 2007; Pearl, 2009). There are two popular frameworks of causal inference: Neyman-Rubin Causal Model (RCM, also known as the potential outcome framework) (Rubin, 1974; Splawa-Neyman et al., 1990) and the Structure Causal Model (SCM) (Pearl, 1995). Within the RCM framework, various models such as Propensity Score Matching (PSM) (Rosenbaum and Rubin, 1983), inverse propensity weighting (IPW) (Rosenbaum, 1987), and Doubly Robust (DR) (Robins et al., 1995) have been employed in transportation fields (Tong and Yu, 2018; Liu et al., 2020; Zhang et al., 2022; Li et al., 2019; Yue et al., 2024). These models are selection-on-observables approaches under the assumption of no unobservable confounders. However, when analyzing traffic crashes and conditions, numerous observable and unobservable confounders may impede the identification of causal effects. There are alternative methods that can account for unobserved confounding bias in the traffic safety field, such as instrumental-variable-based approaches (Afghari et al., 2022), endogeneity models(Bhat et al., 2014), and heterogeneity models(Mannering et al., 2016). Nevertheless, leveraging big data offers an opportunity to mitigate bias from unobservable confounders by providing access to a broader range of variables. In this study, we introduce real-time traffic conditions, road alignment, temporal variables, and other relevant features into the model. This allows us to adjust for potential confounding bias from factors such as weather, special events, light conditions, holidays, etc. (See Section 3 for detailed discussion). Therefore, the selection-on-observables approach is adopted in this study.

Under the RCM framework, this study describes the crash indicator as the "*treatment*" variable and the post-crash speed as the "*outcome*" variable. To facilitate individual-level decision-making in traffic management, it is crucial to consider Heterogeneous Treatment Effects (HTE), which refer to the unique impacts of individual crash on speed. However, previous traffic safety-related studies mainly focused on the causal effects of various factors on crash severity (Afghari et al., 2022; Li et al., 2013; Zhang et al., 2021a, 2022). Very few studies have focused on the causal effect of crashes on speed (Cao et al., 2021; Pasidis, 2019), and their methods are designed for average effects estimation which is not suitable for HTE estimation. In addition, due to the heterogeneity in each road section for each time period, a large number of fixed effects impede the modeling task via traditional regression frameworks. Therefore, in a high-resolution spatiotemporal data environment with various variables, leveraging the modeling and predictive capabilities of ML is essential (Mannering et al., 2020). However, it should be mentioned that the aim of this study is not limited to the sole application of ML models for predicting traffic conditions under crash situations (Grigorev et al., 2022; Moussa et al., 2022; Wen et al., 2022), because these correlation-based models may introduce serious bias when conducting causal inference(Mannering et al., 2020; Prosperi et al., 2020).

Recently, ML has gained significant attention as a replacement for econometric models in the predictive aspect of causal inference. There are various ML-based causal inference methods such as Causal Tree(Athey and Imbens, 2016), Causal Forest(Athey and Wager, 2019), Metalearner(Künzel et al., 2017), Double Machine Learning (DML)(Chernozhukov et al., 2016), Generalized Random Forest (GRF) (Athey et al., 2019) and Doubly Robust Learning (DRL) (Foster and Syrgkanis, 2019). These approaches aim to leverage the power of machine learning algorithms to better handle complex data and capture causal relationships, making them promising tools for causal inference in various fields, including traffic safety research. For instance, Zhang et al. (2021a) used the generalized random forest (GRF) approach to estimate the heterogeneous treatment effect of speed cameras on road safety. Liu et al. (2022) employed the double machine learning (DML) method to estimate the causal effects of curbside pick-up and drop-off on speed. However, there is still a scarcity of research focused on formulating the problem of the causal effect of crashes on traffic conditions using high-resolution data within the framework of causal machine learning, particularly considering varying durations and distances.

In this study, DRL is employed because it maintains many favorable statistical properties (e.g., small mean squared error, asymptotic normality, construction of confidence intervals) and provides a stronger robustness guarantee compared to DML(Foster and Syrgkanis, 2019). However, challenges still need to be addressed: Causal inference requires expert judgment to identify likely confounders, and it is essential to exclude adverse variables from the model (Blakely et al., 2021). Therefore, a suitable method for variable selection is necessary. In addition, unlike ML prediction tasks, causal inference methods lack ground truth for validation since counterfactual data cannot be observed. Therefore, adequate validation approaches are required to evaluate the accuracy and effectiveness of the proposed methods.

To address these challenges, this study introduces a novel causal machine learning framework. This framework not only facilitates causal inference using high-resolution data but also addresses HTE of different types of crashes under various suitations. Ultimately, it provides meaningful causal insights into the impacts of traffic crashes at different upstream locations after different time periods, aiding in informed and accurate decision-making. Our contributions can be summarized as follows:

- The estimation problem of heterogeneous causal effects of crashes is formulated using the RCM. Besides, a causal machine learning framework based on the DRL model is proposed to delve into the intricate ways in which various crashes affect traffic speed under distinct situations.
- A Conditional Shapley Value Index (CSVI) based on causal graph theory is proposed and incorporated into the framework. This addition serves to eliminate adverse null variables and treatment predictors for causal inference, thus enhancing the accuracy of estimated causal effects.
- Multi-source data (e.g., incidents records, traffic flow information, and road alignment data) are merged in a high-resolution way (in 5-minute intervals and 1-mile sections) to estimate the causal effects, and three types of crashes are explored as interventions to understand the causal mechanism at different distances and durations.
- To address the absence of ground truth data for validating the estimated effects, a matching algorithm is proposed to select "counterfactual outcomes" for calculating error metrics. These metrics, along with statistical hypothesis tests and sensitive analyses, are employed to assess the accuracy and robustness of the estimated results.

## 2. Literature Reviews

### 2.1 Traffic crashes impact analysis

Regarding the analysis of the impact of traffic crashes, a considerable amount of research has focused on detecting the impact areas (Pan et al., 2015; Chen et al., 2016; Ou et al., 2020). However, these studies are mainly data-driven and do not provide conclusive evidence on the impact of crashes. Additionally, many scholars have studied the influence of various factors on the severity of crashes(Rolison et al., 2018; Hammond et al., 2019; Zheng et al., 2020; Pu et al., 2020), which play a positive role in understanding the covariates that need to be considered in crashes impact analysis.

This study focuses more on understanding the impact of crashes on traffic speed, while there is limited study analyzing the relationship between traffic crashes and speed directly. Instead, congestion propagation, duration, and delay after crashes are widely analyzed. Chung et al.(2012; 2017) used survival analysis to analyze the effects of different factors on delays and congestion after crashes.

Alder et al. (2013) applied a statistical regression model and estimated the incident duration of non-recurrent congestion. Zheng et al.(2020) identified the determinants of crash-caused congestion by a generalized linear mixed-effects model. With the prevalence of ML, researchers adopted various ML models for predicting the target variables directly. Miller and Gupta(2012) adopted various ML models to predict the cost of delay and incident duration based on calculating economic losses. Xie et al. (2019) built a deep learning model to predict speed after crashes, and a binary classifier was designed to exact the latent impact features for improving performance. Luan et al. (2022) applied a dynamic Bayesian network to predict the congestion of road segments and inferred the propagation. Grigorev et al. (2022) built a bi-level machine learning to predict the duration of incidents with outlier removal, which can deal with the imbalanced data problem.

While there is a wealth of literature on the association between traffic crashes and post-crash outcomes, such as predicting post-crash statements, few studies have focused on the impact of crashes on speed reduction, especially from a causal perspective. Although machine learning models can predict traffic conditions with greater accuracy, their black-box nature does not provide policymakers with a causal explanation of how crashes impact traffic speed.

*2.2 Development of Causal Inference Methods*

Causal inference is a crucial research area in multiple fields, including statistics, medical science, biology, computer science, and economics (Yao et al., 2020). Two popular frameworks for causal inference are RCM and SCM, although they are logically equivalent (Pearl, 2011a). While RCM aims to estimate potential outcomes and treatment effects, SCM mainly defines the causal structure of variables and proposes the probability formula based on the Bayesian network and *do*-calculus to identify causal effects.

In RCM, researchers proposed a propensity score to balance the distributions of covariates. The propensity score indicates the probability of a unit being treated under the given covariates. Propensity score matching (PSM) is the most common approach, which involves matching data from control and experimental groups with identical scores and calculating the difference. However, due to the imbalanced problem, PSM may introduce bias into the estimator. To address this, inverse propensity weighting (IPW) is proposed to re-weight each sample (Rosenbaum, 1987). Furthermore, to solve the bias-variance trade-off dilemma, Robins et al.(1995) proposed a Doubly Robust (DR) estimation method, also named Augmented Orthogonal IPW (AIPW).

Various ML-based causal inference methods have been proposed to deal with heterogeneous treatment effects estimation and improve predictability. Athey and Imbens (2016) proposed Causal Trees by applying "two-tree" estimators to estimate control-outcomes and treatment-outcome conditional on features, and the difference between these estimated outcomes calculates the causal effects. Künzel (2017) further proposed X-learner in which various ML can be applied to estimate outcome and imputed treatment effects (i.e., the treatment effects for individuals in the treated/control group calculated by control-outcome/treatment-outcome estimator), and he also called this model along with the former models as Metalearners. Chernozhukov et al. (2016) introduced Double Machine Learning (DML), which utilizes ML to predict both the outcome and treatment variables. The residuals obtained from this prediction are then used to estimate the heterogeneous treatment effects through ML models. Based on DML methods, Athey et al. (2019) proposed GRF models that mainly utilized a random forest model to balance the covariates, and Foster and Syrgkains proposed DRL(2019) by applying the Doubly Robust form function to estimate the effects with ML models. Various causal machine-learning models have been employed in education, policy, and healthcare(Johansson et al., 2020; Knaus, 2018; Kristjanpoller et al., 2021).

*2.3 Causal Inference in Traffic Safety Analysis*

In traffic safety, most studies mainly applied traditional causal inference methods, such as PSM(Li et al., 2013; Song and Noyce, 2019; Zhang et al., 2021b), DR (Li and Graham, 2016; Graham et al., 2019), to evaluate the effect of management measures to the safety level. In the context of studying the causal effect of crashes on traffic speed, Pasidis (2019) conducted an investigation using a modified differences-in-differences model. Their approach aimed to estimate the average causal effects of traffic crashes on speed. However, their model's design was limited, as it solely focused on calculating the median speed of the same day of the week, time, and location for four weeks before and after the accidents. As a result, this method falls short in satisfying the demand for capturing heterogeneous effects at a fine-grained level. Cao et al. (2021) studied the causal effect of a single crash on traffic speed. They used a dataset with two weeks of traffic speed data, one with the crash and one without it, treating affected road segments as the treatment group and unaffected ones as the control group. They matched control group data to each affected segment using propensity scores and calculated the causal effect as the speed difference between the two groups. However, their study solely concentrated on a single crash, and their method relies on traditional propensity score matching, rendering it less suitable for estimating the HTE of crashes.

Recently, some scholars started paying more attention to causal machine learning methods: Zhang et al. (2021a) applied a GRF model to estimate the road safety level treated by a speed camera. They aggregated the crash data at the site level and combined them with camera site data to estimate the HTE. Liu et al. (2022) estimated the average causal effect of curbside pick-up and drop-offs on speed in different regions based on DML. They considered their model's spatial and temporal features, which improved the performance to capture the congestion effect.

However, few studies focused on the HTE of crashes on traffic speed, especially across varying durations and distance levels. It necessitates the formalization of the problem within a causal inference framework and the handling of large, high-resolution datasets. Besides, although these studies controlled many variables to mitigate the confounding bias, some variables should be excluded from the model to decrease variance and avoid bias (De Luna et al., 2011; Pearl, 2011b).

## 3. Problem Statement

This study explores the causal effect of different types of crashes on highway speed, as congestion levels are primarily linked to crash types (Zheng et al., 2020). The crashes are firstly categorized into rear-end, sideswipe, and crash to objects. Besides, causal inference takes individuals as research objects, but the traffic conditions are continuous. Therefore, the highway are divided into one-mile-long sections and the traffic conditions are aggregated by five-minute intervals to obtain different "individuals" to be estimated.

To identify the causal effects of crashes on speed, we adopt a series stragies to mitigate potential confounding bias. Fistly, we hypothesize that traffic conditions before the crash may influence the occurrence of the crash, while traffic conditions after the crash are a result of the crash's impact. Therefore, we incorporate traffic conditions observed 10 minutes before the crash as confounding variables and the speed after crash as dependent variables. This avoid the endognity problem arising from dependent variable (speed) simultaneously affecting the treatment (crashes). Secondly, the traffic conditions includes traffic flow, speed, occupancy, congestion index, etc. These variables effectively reflect the specific traffic state at that moment, ensuring that the matched control group data is consistent with the traffic conditions before the crash. By controlling these variables, the influence of other potential confounders, such as weather, construction works, and special events, can be reduced. In addition, highways exhibit linear spatial features, and the response of traffic speed to crashes can vary across different sections. To account for this spatial variability, we include each section's milepost and geometric parameters as control variables to effectively fix individual effects. Finally, considering that traffic flow often exhibits apparent temporal dependence, we further incorporate the day of the week and time periods of the day as control variables, which helps to reduce the influence of potential factors such as light conditions and holidays. By controlling the above variables, we believe that the influence of other potential confounding variables can be almost entirely avoided.

We introduce our methodology based on RCM. Let $Y_{s,t}$ denotes the traffic speed in $s$ milepost at time interval $t$, then $Y_{s,t}$ is *potential outcomes*. Let $T_{s,t}^{type} \in \{0,1\}$ be the binary indicator for the *treatment variable* records whether a crash is observed ($T_{s,t}^{type}=1$) or not ($T_{s,t}^{type}=0$), where *type* denotes the type of crash (e.g., rear-end). For further analysis, let $Y_{s,t}^{dur,dis}$ denotes the outcomes under different scenarios, where *dur* denotes the duration since the crash time $t$, and *dis* denotes the upstream distance to crash milepost $s$. Then, based on the RCM framework, the Average Treatment Effect (ATE) can be derived as follows:

$$ATE = E\left[Y_{s,t}^{dur,dis}(1) - Y_{s,t}^{dur,dis}(0)\right]$$
$$Y_{s,t}^{dur,dis} = \begin{cases} Y_{s,t}^{dur,dis}(1) & T_{s,t}^{type}=1 \\ Y_{s,t}^{dur,dis}(0) & T_{s,t}^{type}=0 \end{cases} \quad (1)$$

However, the impact of highway crashes is heterogeneous at different times and locations. For example, as **Figure 1** shows, all these crashes are rear-ended. The crash occurring at morning peak hour caused more serious congestion although at the same place ((1) and (2)), and the crashes occurring at 150 mileposts had little impact on speed than that at 164 mileposts ((1) and (3)). Therefore, we need to introduce covariates $X_{s,t} = \left( x_{s,t}^1, x_{s,t}^2, \cdots, x_{s,t}^k \right)$ and estimate the Conditional Average Treatment Effect (CATE) to explore HTE of crashes. The CATE is defined as a function of full set of conditioning variables to give the conditional mean of crash effect of any point $x_{s,t}^k$, which can be formulated as follows:

$$CATE = E\left[ Y_{s,t}^{dur,dis}(1) - Y_{s,t}^{dur,dis}(0) \mid X_{s,t} \right] \quad (2)$$

Where $X_{s,t}$ is composed of *k*-components, which include features, covariates, or pre-treatment variables that are known not to be affected by crashes (Athey and Imbens, 2016). If $X_{s,t}$ containing all confounders, then we could identify CATE for each subgroup by controlling them. When applying the machine learning, the CATE is conducted for each "individual", thereby facilitating the realization of HTE estimation.

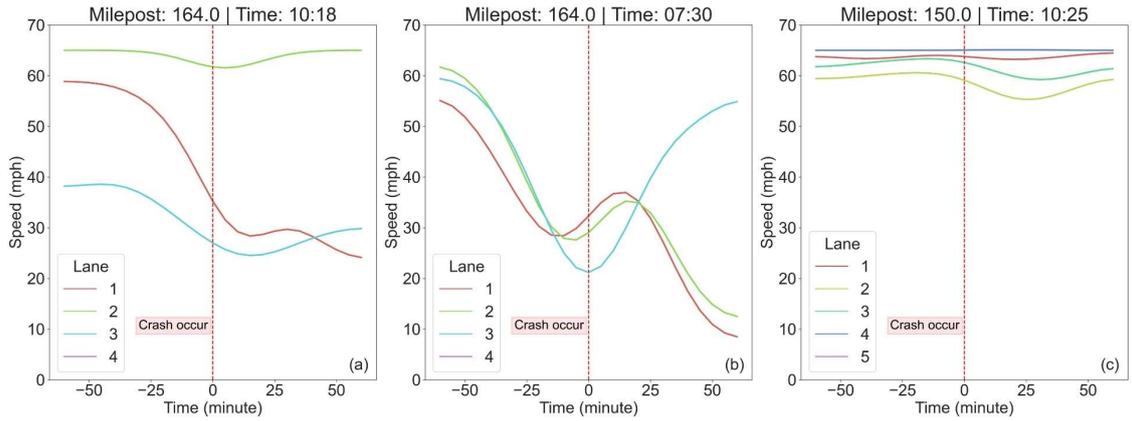

**Figure 1. The change of speed caused by rear-end crashes. (1) and (2) occurred at different periods; (1) and (3) occurred at different locations**

Different from standard traffic condition prediction models, causal effect models cannot be quickly evaluated against each other using a held-out test set because only one of $Y_{s,t}^{dur,dis}(1)$ and $Y_{s,t}^{dur,dis}(0)$ can be observed. Thus, the proper treatment effect itself is never directly observed. This has been termed as the "fundamental problem of causal inference" (Holland, 1985). Therefore, when estimating the causal effect of treatment, the counterfactual outcomes need to be estimated, which will contribute to bias. Therefore, three key assumptions should be satisfied:

***Assumption 1.*** Conditional Independence Assumption (CIA)

$$\left(Y_{s,t}^{dur,dis}(1), Y_{s,t}^{dur,dis}(0)\right) \perp T_{s,t}^{type} \mid X_{s,t} \tag{3}$$

This assumption states that, given the adjustment set $X_{s,t}$, the treatment variable should be independent of the potential outcomes. This assumption is also known as the "no unobserved confounders" assumption, implying that all variables affecting the crash and speed have been controlled. As illustrated above, we incorporate an adequate number of variables to mitigate bias from any potential unobserved confounders.

*Assumption 2.* Stable-unit-treatment-value assumption (SUTVA)

This assumption states that potential outcomes remain unaffected by treatments assigned to other units. However, this can be violated if crashes influence each other or control group traffic states are impacted. To maintain SUTVA, we exclude secondary crashes within 30 minutes of the initial one. We also remove data one hour before/after the crash and within two miles upstream/downstream to prevent treatment influence on control data.

*Assumption 3.* Positivity

$$0 < P\left(T_{s,t}^{type} = 1 \mid X_{s,t}\right) < 1 \quad \text{for all } X_{s,t} \tag{4}$$

This assumption asserts that every combination of covariates in the population has a non-zero probability of receiving the treatment. In other words, there should exist control group data with attributes similar to those of each crash state. However, since the quantity of crash data is significantly less than normal data, random sampling of control data may not satisfy this requirement. To address this, we select normal data with matching period, location, and direction as control, ensuring reasonable satisfaction of the positivity assumption.

## 4. Methodology
### *4.1 Traffic Crash Causal Effect Analytic Framework*

In this paper, we propose a causal analytic framework that adopts machine learning causal inference models for inferring the spatiotemporal causal effect of different types of traffic crashes on traffic speed based on contributing features. As illustrated in **Figure 2**, firstly, we merge traffic flow, crash, and road alignment data. The crash and non-crash data and the corresponding features are divided into experimental and control groups. The outcome Y, treatment T, and features X can be determined based on the target scenarios. Then, we employ directed acyclic graph (DAG) theory to select variables and remove collinearity. Thirdly, based on the SCM, we established the causal relationship of the selected variables, treatment, and outcome. At last, we chose the most suitable machine learning model for classification and regression via grid search cross-validation and introduced the data and models into the DRL model to estimate the causal effect of different crash types on different duration and distances. Our framework incorporates causal inference and ML into traffic crash analysis, which could serve as a valuable reference for future research in this area.

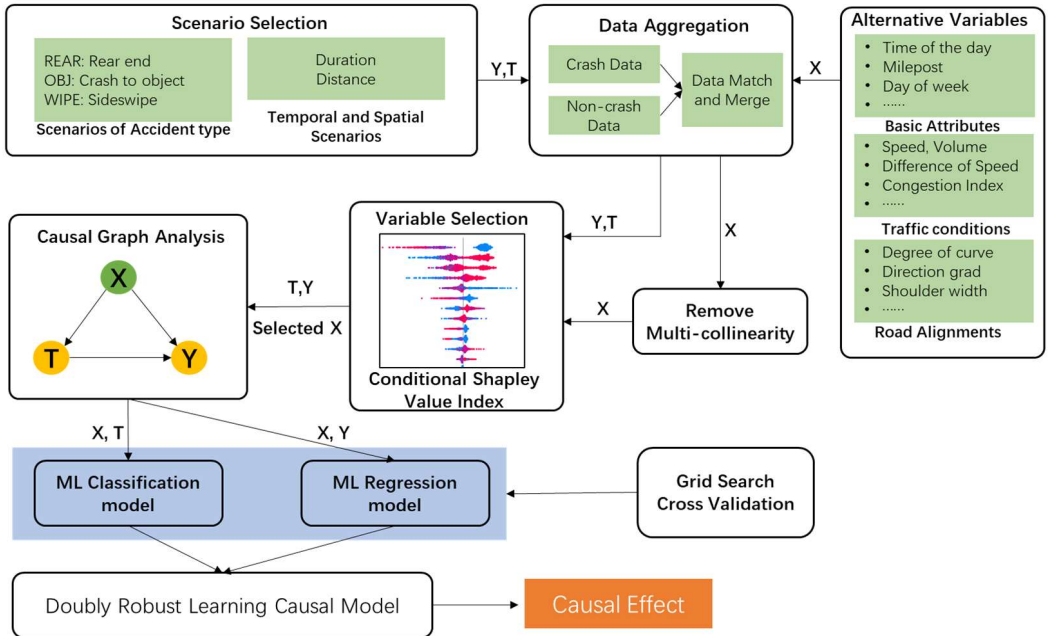

**Figure 2. The framework of highway crashes causal inference**

*4.2 Variables selection based on Shapley Value*

The primary purpose of variable selection is to reduce bias and improve efficiency for causal effect estimation(Witte et al., 2019). Although there are no straightforward methods to ensure the best variable set, we can develop some basic strategies based on causal graph theory to select variables.

*4.2.1 Basic variables selection criteria*

The causal graph is widely used in SCM, reflecting the qualitative causal mechanism of factors. The DAG represents this graph and can be determined according to prior knowledge and data. According to the causal graph theory, we could divide the covariates into four categories $X = \{X_C, X_T, X_O, X_N\}$ (Tang et al., 2020) and define $X_C$ as confounders that have direct causal effects both on speed and crahes. $X_O$ and $X_T$ are used to denote outcome predictors and treatment predictors, respectively. These terms refer to the covariates that exclusively affect speed or crashes causally. $X_N$ denotes as null variables, which represents variables other than the above three types of variables. The corresponding DAG diagram is shown in **Figure 3**.

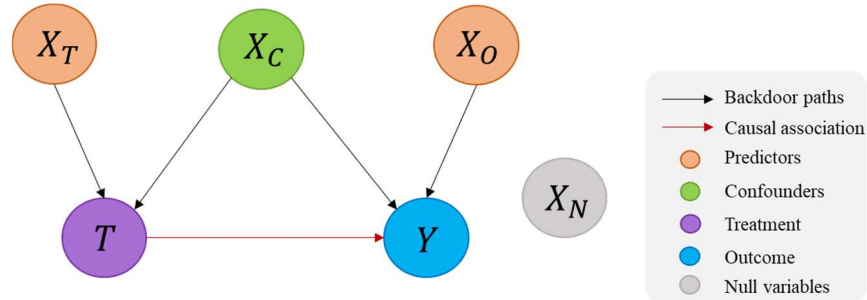

**Figure 3. The causal mechanism of variables demonstrated by the directed acyclic graph**

Among these variables, including all confounders is the key to correctly identifying the treatment effect. However, even though the outcome and treatment still render the conditional independent requirement, the inclusion of treatment predictors into the conditioning sets may contribute to the increased variance of the estimated treatment effect (Brookhart et al., 2006; Rotnitzky and Smucler, 2020; Schnitzer et al., 2016). Moreover, studies show that this inclusion may cause bias (De Luna et al., 2011; Pearl, 2011b). Besides, null variables may function as colliders which should not be controlled in causal inference because conditioning on colliders may induce association between the covariates and contribute to estimated bias.

*4.2.2 Remove multi-collinearity and adverse variables*

In causal modeling, it's not uncommon for two or more variables to exhibit a strong correlation, often manifesting as a linear relationship. This deterministic interdependency, known as collinearity, can pose challenges in data analysis(Spirtes and Scheines, 2004). Apart from the demanding task of accurately specifying causal models, bias can be exacerbated with increasing collinearity (Schisterman et al., 2017). The parameters related to traffic conditions often display significant correlations, such as volume and speed. To address this, we firstly employed the Pearson correlation coefficient to quantify the linear associations between variables, subsequently removing covariates with substantial linear relationships.

Remain covariates may still include adverse variables (i.e. treatment predictor and null variables). According to the causal graph in **Figure 3**, if the covariate has no correlation with outcome conditioned on treatment, they can be removed. To test the conditional correlation, Shapley value is adopted to test the correlation between remained covariates and speed conditioned on crash. SHAP is an interpretation method that utilizes the Shapley value based on game theory to combine optimal credit allocation with local explanations (Lundberg and Lee, 2017). Let $Shap(X(t), Y(t))$ denote the absolute Shapley value of variable $X$ for speed $Y$ under crash treatment $T = t$. If $Shap(X(t), Y(t)) = 0$, we consider $X$ is independent of $Y$ when $T = t$.

We propose the Conditional Shapley Value Index (CSVI) to perform conditional independence. Let $\omega = P(T = 1)$ be the probability of receiving a crash, and the CSVI between $X$ and $Y$ given $T$ is defined as follows:

$$CSVI(X, Y | T) = \omega Shap(X(1), Y(1)) + (1 - \omega) Shap(X(0), Y(0)) \qquad (5)$$

This index quantifies the importance of variables $X$ for $Y$ when the crash occurs or not. According to the definition of treatment predictors and null variables, if the variable $X$ is not related to $Y$ no matter the value of $T$, then $Shap(X(1), Y(1)) = Shap(X(0), Y(0)) = 0$. Considering the collected traffic data are not balanced, we add a weight to balance the index. Let $\hat{\omega} = n_t / n = \sum_i^n T_i(1) / n$ be the empirical estimator of $\omega$, where $n$ is the number of all samples and $n_t$ is the number of crash samples. The closer the $CSVI(X, Y | T)$ is to zero, the more likely $X \perp Y | T$, which means the $X$ belongs to $X_T$ or $X_N$ that should be excluded.

***Principle 1.*** If $CSVI(X, Y | T) < \varepsilon$, the $X$ should be ignored, where $\varepsilon$ is a threshold determined according to the experiments.

To determine the threshold $\varepsilon$, we can apply a series of thresholds to estimate the causal effects and select the best value based on the estimated performance. The validation method to evaluate the performance will be discussed lately (Section 4.5).

*4.3 Causal Graph Analysis of Traffic Crashes*

After the variable selection, we assume there are no adverse variables, and the outcome predictors and confounders can be controlled when estimating the CATE. We determine the controlled variables based on the SCM theory to understand the causal relationship between the variables further. To simplify the illustration, the confounders are divided into three categories: 1) basic attributes $X_{s,t}^{Basic}$, such as period, milepost, direction, and city; 2) road alignment $X_{s,t}^{Align}$, referring to the geometric parameters of each road segment, such as gradient and degree of the curve; 3) traffic preconditions $X_{s,t}^{Condi}$, referring to the traffic conditions 10 minutes before the crash occurrence. The causal graph of these three types of covariates, crash treatment $T_{s,t}^{type}$, and traffic speed $Y_{s,t}^{dur,dis}$ are shown in **Figure 4**.

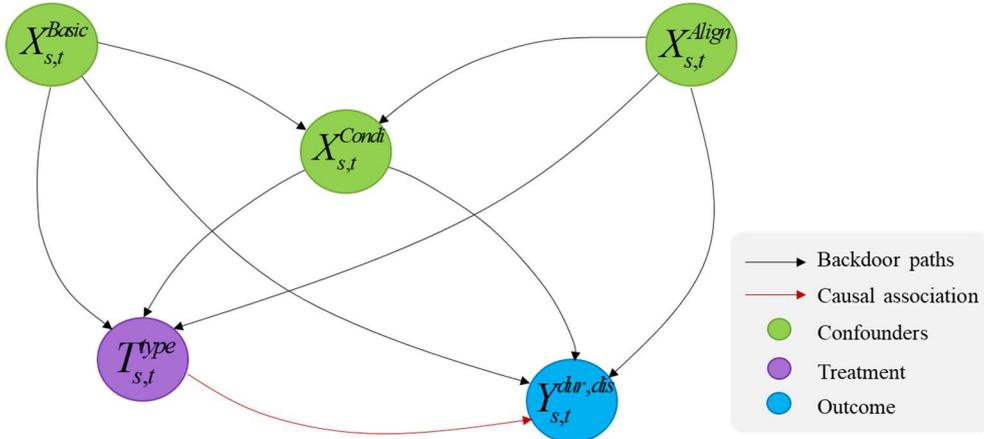

**Figure 4. The causal graph of the covariate, traffic crash, and speed**

We demonstrate three categories of variables to represent all factors for ease of display, and we claim that all the factors belonging to the corresponding category satisfy the relationship in the diagram. According to the causal graph, the red arrow refers to the causal association of traffic crashes to the traffic speed, which is our estimation target. The paths $T^{type} \leftarrow X \rightarrow Y^{dur,dis}$ are *backdoor paths*, which should be blocked during estimation. Therefore, the $X^{Basic}$, $X^{Align}$ and $X^{Condi}$ should be controlled during estimation. Furthermore, in order to estimate the causal effect of crashes on traffic speed at various upstream locations after different periods, we assume as follows:

***Assumption 4***: For any traffic statement, no additional factor affects the traffic speed $dur$ minutes after the crash and $dis$ miles to the crash location.

Therefore, based on ***Assumption 4***, formulate the statistical estimand of conditional average traffic crash causal effect for different duration and distances in Equation (6).

$$\widehat{CATE}_{dur,dis}^{type} = E_{X_{s,t}} \left[ E\left(Y_{s,t}^{dur,dis}\left(T_{s,t}^{type}=1\right) | X_{s,t}\right) - E\left(Y_{s,t}^{dur,dis}\left(T_{s,t}^{type}=0\right) | X_{s,t}\right) \right] \quad (6)$$

Where $X_{s,t} = \left(X_{s,t}^{Basic} \cup X_{s,t}^{Aling} \cup X_{s,t}^{Condi}\right)$.

### *4.4 Doubly Robust Learning Causal Inference*

To improve the robustness of inference, we apply doubly robust learning methods to estimate the average treatment effect of crashes. As introduced before, the DR estimator is formulated based on the propensity score, defined as the probability of a crash to the traffic state conditional on observed variables (Equation 7).

$$e(x) = P(T=1 | X=x) \quad (7)$$

The traffic states with the same propensity score are recognized as conditional independence, and their speed from the experiment and control group are the potential outcomes. Thus, the average treatment effect can be calculated as Equation 8, and this process is known as *matching*.

$$\widehat{CATE} = E\left(Y_{s,t}(1) | e(x_{s,t}) = \gamma\right) - E\left(Y_{s,t}(0) | e(x_{s,t}) = \gamma\right), \; \gamma \in (0,1) \quad (8)$$

To solve the problem of imbalanced data, inverse propensity weighting (IPW) is applied to re-weight each sample as:

$$r = \frac{T}{e(x)} - \frac{1-T}{1-e(x)} \quad (9)$$

Therefore, the re-weighted IPW estimator of CATE can be formulated as follows:

$$\widehat{CATE}_{IPW} = E\left(\frac{T_{s,t} Y_{s,t}}{\hat{e}(x_{s,t})}\right) - E\left(\frac{(1-T_{s,t}) Y_{s,t}}{1-\hat{e}(x_{s,t})}\right) \quad (10)$$

Furthermore, the DR estimator combines the IPW and a regression model to solve the bias-variance problem. As long as one is correctly specified, the estimator can correctly estimate CATE, which is *doubly robust*. According to our problem, the estimator can be formalized as follows:

$$\widehat{CATE}_{DR,dur,dis}^{type} = E\left[ Y_{DR,s,t}^{dur,dis}(1) - Y_{DR,s,t}^{dur,dis}(0) \mid X_{s,t} \right]$$

$$Y_{DR,s,t}^{dur,dis}(1) = \frac{T_{s,t}^{type} Y_{s,t}^{dur,dis}}{\hat{e}^{type}(X_{s,t})} - \frac{T_{s,t}^{type} - \hat{e}^{type}(X_{s,t})}{\hat{e}^{type}(X_{s,t})} \hat{m}^{dur,dis}\left(T_{s,t}^{type}=1, X_{s,t}\right) \quad (11)$$

$$Y_{DR,s,t}^{dur,dis}(0) = \frac{(1-T_{s,t}^{type}) Y_{s,t}^{dur,dis}}{1-\hat{e}^{type}(X_{s,t})} - \frac{T_{s,t}^{type} - \hat{e}^{type}(X_{s,t})}{1-\hat{e}^{type}(X_{s,t})} \hat{m}^{dur,dis}\left(T_{s,t}^{type}=0, X_{s,t}\right)$$

Where $\hat{e}^{type}(X_{s,t})$ is the propensity score estimated model for indicator variable of $T_{s,t}^{type}$, $\hat{m}^{dur,dis}(T_{s,t}^{type}, X_{s,t})$ is the regression model for the traffic speed $Y_{s,t}^{dur,dis}$ based on the covariates $T_{s,t}^{type}$ and $X_{s,t}$.

The DRL model follows the two-stage process, where ML models estimate the propensity score and outcome in the first stage, and the CATE is estimated in the second stage. In the first stage, based on the causal graph analysis in Section 4.3, we formulate the functions to predict $\hat{m}^{dur,dis}(T_{s,t}^{type}, X_{s,t})$ and $\hat{e}^{type}(X_{s,t})$ as follows:

$$\hat{m}^{dur,dis}(T_{s,t}^{type}, X_{s,t}) = \hat{Y}_{s,t}^{dur,dis} = \hat{g}_{type}^{dur,dis}\left(T_{s,t}^{type}, X_{s,t}^{Basic}, X_{s,t}^{Condi}, X_{s,t}^{Align}\right) + \varepsilon_{s,t}^{dur,dis} \quad (12)$$

$$\hat{e}^{type}(X_{s,t}) = \Pr(T_{s,t}^{type} \mid X_{s,t}) = \hat{p}^{type}\left(X_{s,t}^{Basic}, X_{s,t}^{Condi}, X_{s,t}^{Align}\right) + \delta^{type} \quad (13)$$

Where $\hat{g}_{type}^{dur,dis}$ predicts the value of $Y_{s,t}^{dur,dis}$ under the crash $treat$, $\hat{p}^{type}$ predicts the probability of $T_{s,t}^{type} \in \{0,1\}$, $\varepsilon_{s,t}^{dur,dis}$ and $\delta^{type}$ are unobservable noise with $E(\varepsilon_{type}^{dur,dis} \mid X_i) = 0$ and $E(\delta^{type} \mid X_i) = 0$.

Considering $T_i^{treat}$ is a binary variable and $Y_i^{dur,dis}$ is a continuous variable, we apply classification ML models and regression ML models to predict them, respectively.

In the second stage, it should be noted that the true CATE cannot be obtained since the counterfactual outcomes are predicted based on observational data. Thus, it is not necessary to pay much attention to the accuracy of the final model since the input variables are estimated. This study applies Ordinary Least Square (OLS) to estimate the $\widehat{CATE}_{DR,dur,dis}^{type}$ according to Equation (11) by regressing $\hat{Y}_{DR,s,t}^{dur,dis}(1) - \hat{Y}_{DR,s,t}^{dur,dis}(0)$ on $X_{s,t}$. After fitting the causal machine learning model, the heterogeneous treatment effect can also be estimated when inputing the data of each "individual".

*4.5 Validation based on Matching Algorithm*

Because the counterfactual outcomes cannot be observed, there is no ground truth data to evaluate the estimated errors. Therefore, we propose a matching algorithm to search the reference data that functioned as "counterfactual outcomes" for verification. Due to the inherent periodicity in traffic flow, it is readily feasible to identify numerous non-crash traffic conditions that exhibit spatiotemporal consistency with crash states. As such, these conditions serve as a reference group for comparative analysis. Pasidis (2019) applied matched data as counterfactuals in the control group, demonstrating the validity of using matched data for estimating the causal effects of crashes. Differing from his study, matching data is not applied for the control group during the modeling task but for validation. The average non-crash conditions for all locations and time periods are input into the model for training, therefore, the matched data for validation does not overlap with training data to a certain extent.

During the matching process, several non-crash conditions are selected for each crash. To be exact, for a given crash data $c_i$ from experimental data group $S_{expri}$, we search the corresponding $k$ non-crash traffic states $N_i = \{n_1, n_2, \cdots, n_k\}$ from the control group $S_{ctrl}$ based on the *time, week milepost,* and *direction* of $c_i$. Each data $c_i$ or $n_i$ is a vector data containing all variables (such as pre-treatment traffic conditions) and outcomes. Algorithm 1 shows the procedure of "counterfactual outcomes" matching.

---

**Algorithm 1** Data Matching: match "counterfactual outcomes" for crash conditions

---

**Input:** $S_{expri}$: dataset of crash data; $S_{ctrl}$: dataset of non-crash data; *TimePeriod* : a given time range for search; *K*: number of expected matched data
**Output:** $\mathbb{N}$ : the matched non-crash dataset, $\mathbb{N} = \{N_1, N_2, \cdots N_i\}$

---

$\mathbb{N} = \{\}$
**For** $c_i$ **in** $S_{expri}$ **do**
   $t = c_i.time$ , $w = c_i.week$ , $mp = c_i.milepost$ , $dir = c_i.direction$
   $k = 0$, $N_i = []$
  **For** $n_j$ **in** $S_{ctrl}$ **do**
    **If** $n_j.time$ **in** Range(*Timeperiod*) **then**

    **If** $t == n_j.time$ **and** $w == n_j.week$ **and** $mp == n_j.milepost$ **and** $dir == n_j.direction$

      $N_i.add(n_j)$, $k \mathrel{+}= 1$

    **If** $k == K$ **then**  // Avoid spending much time matching data

  **End for**

  $\mathbb{N}.add(N_i)$

**End for**

For each crash $c_i$, we define the original speed as $osp_i$ and estimated individual treatment effect is $ice_i$. After we matched the non-crash dataset $N_i$, the "ground truth" speeds $SP_i = [sp_1, sp_2, \cdots, sp_k]$ can be obtained, and the matched causal effects are:

$$mce_i = \sum_k (csp_i - sp_k)/k \tag{14}$$

Where $csp_i$ is the post-crash speed of $c_i$ calculated by $osp_i + ice_i$. Based on these matched causal effects, we can calculate the performance of estimation, which will be discussed in Section 5.5.2.

It should be noted that different from the before-after analysis that compares historical observational data to calculate the effects, our method is to build a model that can be applied to predicting HTE. Therefore, this method exhibits superior efficacy in simultaneously analyzing diverse crashes and accurately estimating heterogeneous treatment effects.

## 5. Experiments and Results

### 5.1 Data Source and Preparation

The main road of Interstate 5 (I5) in Washington is selected as our research object. The data used in this study range from Nov. 2019 to Jun. 2021, and the milepost range from 139 to 178, consisting of three data sets:

**Traffic data set**: The traffic data record the traffic conditions by three parameters detected by the loops: Volume (vol), Occupancy (occ), and Speed (spd), and also provide information including time, location (milepost), road type, direction and the number of lanes.

The raw data is recorded by detectors installed unevenly on the road. We pre-process this data to obtain traffic conditions at each milepost using two rules: 1) Calculate the mean values of the speed and occupancy and the sum of the volume recorded by the detectors within one mile. 2) Interpolate the nearest two data points for unrecorded mileposts. Furthermore, all data is aggregated in 5-minute intervals.

**Incident data set**: The incident data are collected by Highway Safety Information System (HSIS), focusing on the 4815 crashes within the scope of research. This data contains basic information about the crashes (e.g., time, location, weather), collision type (e.g., sideswipe, rear-end), and subtypes (e.g., vehicle type, driver status). We match the crash data with the traffic data based on location and time.

**Road Alignment data set**: The alignment data comes from roadway inventory files collected by HSIS and contain three subfiles (Roadlog, Curve, and Grade) "section" files. Each homogeneous roadway section is defined by a beginning and ending milepost. The basic Roadlog file contains information like lane, shoulder width, and type, while the supplemental files provide horizontal and vertical alignment information. To match traffic data, we rearranged the data as the "point data" so that the alignment characteristics are described at a given milepost.

We chose one and two miles upstream and downstream of the given milepost as the related positions and used their six parameters 10 minutes before a given time. The variables set is outlined as follows:

**Table 1 The description of candidate variables.**

| Category | Variables | Values | Description |
|---|---|---|---|
| Basic attributes | *time* | 0: night, 1: off-peak, 1: peak hour | The period during the day, |
| | *milepost* | [139,178] | The milepost of the road |
| | *direction* | 0: Northward, 1: Southward | The direction of the road |
| | *week* | {0, 1, …, 6} | The day of the week |
| | *aadt* | $\mathbb{R}^*$ | Average annual daily traffic |
| | *city* | $\mathbb{R}^*$ | City number |
| | *pop_grp* | $\mathbb{R}^*$ | City population |
| Traffic conditions* | *vol* | $\mathbb{R}^*$ | The volume of vehicles |
| | *occ* | $\mathbb{R}^*$ | The occupancy time |
| | *spd* | $\mathbb{R}^*$ | The average speed during the time interval within 1 milepost |
| | *std_spd* | $\mathbb{R}^*$ | The standard deviation of speed |
| | *spd_diff* | $\mathbb{R}^*$ | The max speed difference between lanes |
| | *ci* | [0,1] | Congestion Index |
| Road Alignment Features | *lanewid* | $\mathbb{R}^*$ | The width of the lane |
| | *medwid* | $\mathbb{R}^*$ | The width of median |
| | *shlwid* | $\mathbb{R}^*$ | The width of the shoulder |
| | *no_lane* | {2, 3, 4, 5} | The number of lanes |
| | *curv_max* | $\mathbb{R}^*$ | The max degree of the curve |
| | *deg_curv* | $\mathbb{R}^*$ | The degree of the curve |
| | *pct_grad* | $\mathbb{R}^*$ | The percentage of gradient |
| | *dir_grad* | 0: up, 1: down | The direction of the gradient |

*Note: All variables of traffic conditions includes the variables upstream and downstream to the occurrence place and are denoted in models with suffix down_1, down_2, up_1, and up_2.*

*5.2 Traffic Crash Scenarios Definition and Characteristics*

After preliminary analyzing the types of the crash in the data set, the targets of traffic crashes in this study are classified into three main types: Rear-end (REAR), Crash to object (OBJ), and Sideswipe (WIPE), and the sample sizes of each type are 1959, 713, and 922, respectively, after data cleaning (incident types with less than ten occurrences are deleted).

To explore the spatiotemporal causal effects of crashes, we estimated the causal effect 5 minutes to 30 minutes after the crash happened from the place of the crash to the 5 miles upstream. Therefore, we can define that $dur \in [5, 10, 15, 20, 25, 30]$, $dis \in [0, -1, -2, -3, -5]$, $type \in [REAR, OBJ, WIPE]$. Here, the values in $dis$ denote the relative distance to the place where the crash occurred.

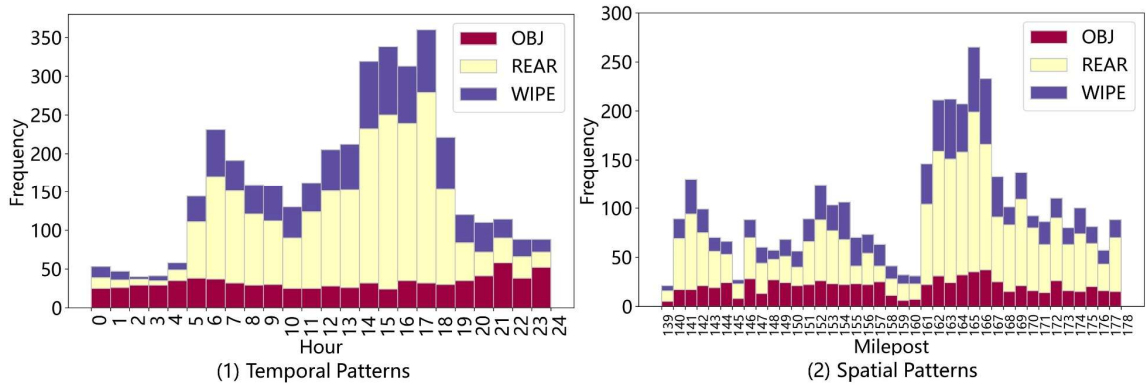

**Figure 5. The spatial and temporal distribution of different crash types**

As shown in **Figure 5**, The spatial distribution of crashes depends on whether the road passes through an urban area. There is a particular section in Seattle, from 162 to 167mp, where crashes are more likely to occur. REAR and WIPE occur more often during the day, especially during peak hours, while OBJ crashes happen more frequently at night due to poor visibility and driver fatigue.

According to the data in **Figure 6**, each type of crash has a unique effect on traffic conditions compared to non-crash data. REAR crashes result in the highest means (red dots) of occupancy and volume, leading to the lowest means of traffic speed. WIPE has a similar effect but less impact on traffic conditions. Traffic conditions associated with OBJ crashes exhibit greater similarity to non-crash conditions. Particularly noteworthy are the boxplots illustrating the lane-based speed difference and congestion index for OBJ and non-crash scenarios, both of which are centered around zero. This pattern stands in stark contrast to the other two crash types. Consequently, these two parameters, namely *spd_diff* and *ci*, hold potential utility for models aimed at distinguishing between different crash types.

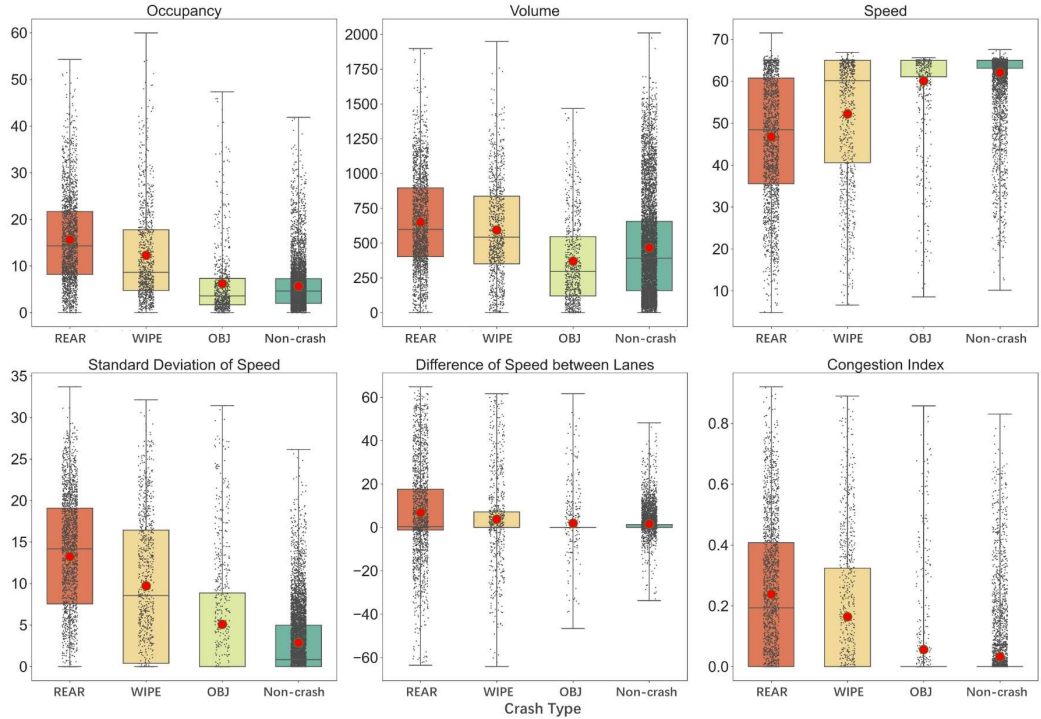

**Figure 6. The statistic distributions of basic traffic parameters based on the types (red dots are mean values)**

*5.3 Variable Selection Results*

We conducted a test for multicollinearity using the Pearson correlation coefficient. The results are displayed in **Figure 7**, where each square represents a coefficient between two variables, with colors corresponding to the coefficient values as indicated by the color bar legend. We observed a high correlation between occupancy and speed, as well as between the standard deviation of speed and the congestion index. Consequently, we retained only one of these correlated variables in the model. We chose *ci* since it has a lower coefficient with *vol* than the other three and shows higher differentiation for different types of crashes (**Figure 6**). We removed *occ*, *spd*, and *std_spd* from the model. Additionally, we deleted *medwid* and *pct_grad* since they have high correlations to *milepost* and *dir_gard*, respectively.

Considering the large computational burden of our experiments, we use Light Gradient Boosted Machines (LGBM) to calculate the Shapley values due to its implementation of two techniques: Gradient-Based One-Side Sampling and Exclusive Feature Bundling. These techniques enable faster execution and higher accuracy(Ke et al., 2017). **Figure 8** summarizes the effects of all features. The horizontal axis represents the SHAP values, and the left vertical axis shows the 20 most critical variables, sorted by their mean absolute Shapley value, and the right vertical axis shows the value of the feature. The redder the color, the higher the value of the feature, and vice versa. For instance, for the control group of the REAR crash, the *ci_up1* has the highest mean absolute SHAP value, indicating its significant contribution to predicting speed, and a higher *ci_up1* leads to a lower *speed*. **Figure 8** reveals that the variables of traffic conditions contribute significantly to the outcome, and the difference in SHAP values between the control and experiment groups is apparent. To eliminate bias and variance, we calculated the CSVI for each variable (**Figure 9**) and removed variables such as *no_lane*, *lanewid*, *city*, *dir_grad*, *pop_grp*, *aadt*, and *curv_max*, based on Principle 1, with a threshold of $\varepsilon=0.15$.

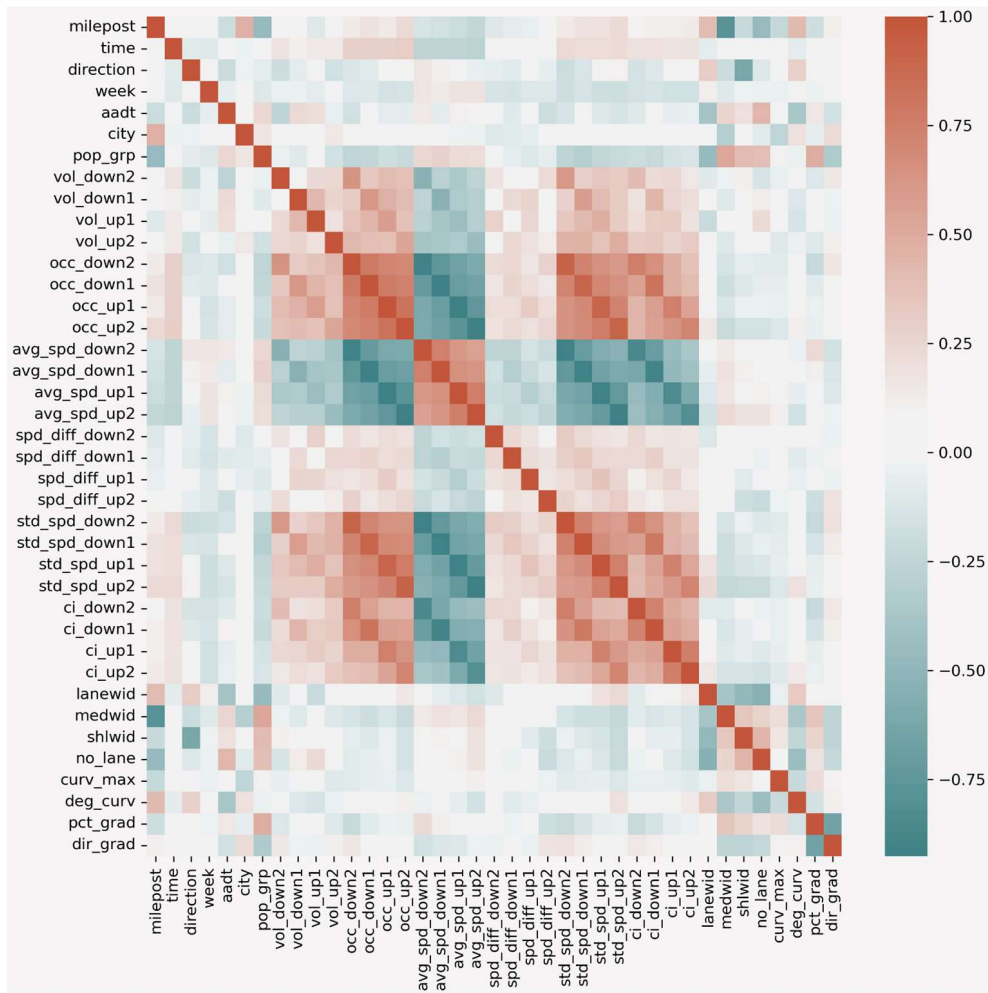

**Figure 7. The Pearson correlation coefficient of all candidate variables.**

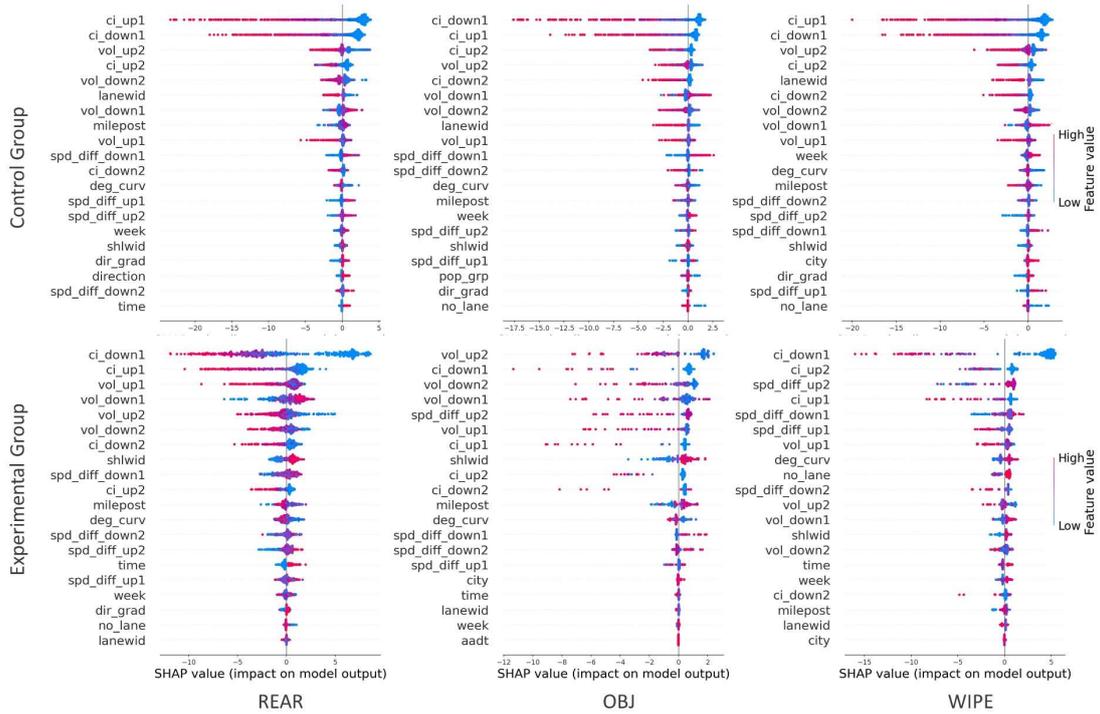

**Figure 8. Shapley values of the variables for the speed in experimental and control group.**

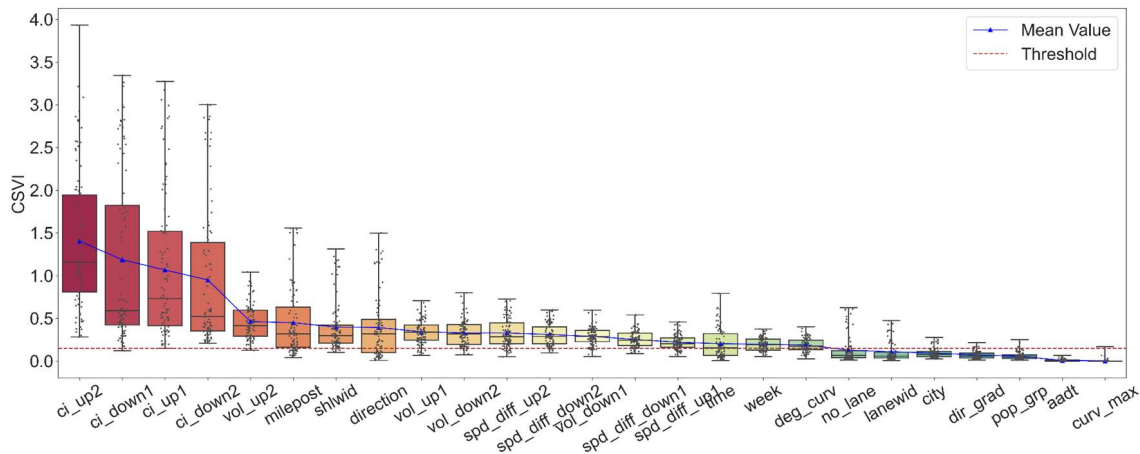

**Figure 9. Distribution of *CSVI* results for each variable**

*5.4 Classification and Regression Machine Learning Estimation Results*

We conduct specific screening for the ML models in the first stage. We take several alternative ML models for classification and regression prediction, and the alternative models are listed as follows:1) extreme gradient boosting (XGBoost)(Chen and Guestrin, 2016),2) Random Forest (RF)(Breiman, 2001), 3) Light gradient boosted machines (LGBM)(Ke et al., 2017),4) Linear Regression (LR) and 5) Support Vector Machine (SVM).

**Table 2. The searched hyperparameters of each model.**

| Model | Hyperparameters | Search Range |
|---|---|---|
| XGBoost | n_estimators | [10, 40, 60, 80, 100] |

|  | max_depth | [3, 4, 5, 6, 7, 8, 9, 10, None] |
|  | learning_rate | [0.01, 0.02, 0.03, 0.04, 0.05] |
|  | subsample | [0.5, 0.6, 0.7, 0.8, 0.9, 1.0] |
|  | min_child_weight | [1, 2, 3, 4, 5, 6, 7, 8, 9, 10] |
| LGBM | n_estimators | [10, 40, 60, 80, 100] |
|  | max_depth | [3, 4, 5, 6, 7, 8, 9, 10, -1] |
|  | learning_rate | [0.01, 0.02, 0.03, 0.04, 0.05] |
|  | subsample | [0.5, 0.6, 0.7, 0.8, 0.9, 1.0] |
|  | min_child_samples | [10, 20, 30, 40, 50] |
|  | min_child_weight | [1e-3, 1e-2, 1e-1, 1, 1e1, 1e2] |
|  | max_bin | [255, 510, 765, 1020] |
| RF. | n_estimators | [10, 15, 20, 25, 30] |
|  | max_depth | [3, 5, 7, 9, 11, None] |
|  | min_samples_split | [2, 3, 4, 5] |
|  | min_samples_leaf | [1, 2, 3, 4, 5] |
|  | max_features | ['auto', 'sqrt', 'log2', None] |
| LR. | fit_intercept | [True, False] |
|  | normalize | [True, False] |
|  | n_jobs | [-1, 1, 2, 4, 8] |
|  | positive | [True, False] |
| SVM | C | [0.001, 0.01, 0.1, 1, 10, 100] |
|  | kernel | ['linear', 'poly', 'rbf', 'sigmoid'] |
|  | degree | [1, 2, 3, 4, 5, 6] |
|  | gamma | ['scale', 'auto', 0.01, 0.1, 1, 10, 100] |

To select the best models, we must choose a series of hyper-parameters to train and test the data, and the searched hyperparameters are outlined in Table 2. Considering large quantities of parameters in ML models, especially XGBoost and LGBM, grid searching for all parameters will consume much time. Thus random grid search was applied, which could randomly choose hyper-parameters more efficiently within a small fraction of the computation time(Bergstra and Bengio, 2012). After grid searching, the best parameters were input into the model for cross-validation. In this study, *k*-fold cross-validation was employed, enabling all observations to be used both for training and testing by splitting the sample data into *k* equal-sized subsamples. We applied 10-fold cross-validation for model training, dividing the data into ten equal subsets. Among the ten subsamples, one functions as the validation data, and the remaining nine subsamples were used as the training data. To select the best models, several indicators for classification and regression models were applied to evaluate the prediction accuracy. For classification, Precision, Accuracy, Recall, and F1-score are used to quantify the error. These indicators are determined as follows:

$$presicion = \frac{TP}{TP+FP} \quad (15)$$

$$accuracy = \frac{TP+TN}{TP+TN+FP+FN} \tag{16}$$

$$recall = \frac{TP}{TP+FN} \tag{17}$$

$$F1 = \frac{2TP}{2TP+FN+FP} \tag{18}$$

Where $TP, TN, FP, FN$ denote the true positives, true negatives, false positives, and false negatives, respectively.

As for regression, we used Mean Absolute Error (MAE), Mean Squared Error (MSE), Root Mean Squared Error (RMSE), Mean Absolute Percentage Error (MAPE), and R-squared ($R^2$). The formulas are shown as follows:

$$MAE = \frac{1}{n}\sum_{i}^{n}\left|\hat{y}_i - y_i\right| \tag{19}$$

$$MSE = \frac{1}{n}\sum_{i}^{n}\left(\hat{y}_i - y_i\right)^2 \tag{20}$$

$$RMSE = \sqrt{\frac{1}{n}\sum_{i}^{n}\left(\hat{y}_i - y_i\right)^2} \tag{21}$$

$$MAPE = \frac{1}{n}\sum_{i}^{n}\left|\frac{\hat{y}_i - y_i}{y_i}\right| \tag{22}$$

$$R^2 = 1 - \frac{\sum_{i}^{n}\left(\hat{y}_i - y_i\right)^2}{\sum_{i}^{n}\left(y_i - \bar{y}\right)^2} \tag{23}$$

Where $y_i$ denotes the actual value of the target variable, $\hat{y}_i$ denotes the corresponding predicted value, and $\bar{y}$ denotes the average value of the target variable.

Table 3 and Table 4 present the performance results. The Random Forest model is best suited for classification and regression, even though sometimes LGBM performs better in specific error parameters. Therefore, we chose the Random Forest model as the base ML model to predict propensity scores and outcomes.

Table 3. Performance results of classification for different types of crash type

| Model | Criterion | XGBoost | LGBM | RF. | SVM | Best Model |
|---|---|---|---|---|---|---|
| REAR | Accuracy | 0.9688 | 0.9692 | **0.9704** | 0.5363 | RF. |
| | Recall | 0.9577 | 0.9552 | **0.9580** | 0.5414 | |
| | Precision | 0.9786 | **0.9820** | 0.9817 | 0.5212 | |
| | F1 | 0.9679 | 0.9683 | **0.9696** | 0.4483 | |
| OBJ | Accuracy | 0.9930 | 0.9932 | **0.9938** | 0.6317 | RF. |
| | Recall | 0.9760 | **0.9771** | 0.9765 | 0.3348 | |
| | Precision | 0.9970 | 0.9967 | **0.9996** | 0.2671 | |
| | F1 | 0.9862 | 0.9867 | **0.9878** | 0.2313 | |
| WIPE | Accuracy | 0.9740 | 0.9755 | **0.9770** | 0.5752 | RF. |
| | Recall | 0.9341 | **0.9372** | 0.9349 | 0.3327 | |
| | Precision | 0.9822 | 0.9841 | **0.9912** | 0.2726 | |
| | F1 | 0.9572 | 0.9597 | **0.9619** | 0.2215 | |

Table 4. Performance results of regression for different types of crash type

| Model | Criterion | XGBoost | LGBM | RF. | LR. | Best Model |
|---|---|---|---|---|---|---|
| REAR | MAE | 3.3800 | 3.2483 | **3.1836** | 4.8665 | RF. |
| | MSE | 46.3607 | **42.2842** | 42.3343 | 65.8335 | |
| | RMSE | 6.7393 | **6.4333** | 6.4389 | 8.0456 | |
| | R2 | 0.6403 | 0.6725 | **0.6734** | 0.4757 | |
| | MAPE | 0.1021 | 0.1004 | **0.0998** | 0.1425 | |
| OBJ | MAE | 1.6401 | 1.6005 | **1.5504** | 2.5869 | RF. |
| | MSE | 17.3722 | 15.9016 | **15.8896** | 26.2485 | |
| | RMSE | 4.0902 | 3.9126 | **3.9103** | 5.0345 | |
| | R2 | 0.6981 | 0.7241 | **0.7244** | 0.5309 | |
| | MAPE | 0.0409 | 0.0407 | **0.0402** | 0.0616 | |
| WIPE | MAE | 2.2389 | 2.1737 | **2.1321** | 3.3788 | RF. |
| | MSE | 25.7684 | **23.4542** | 23.8749 | 37.7180 | |
| | RMSE | 5.0087 | **4.7752** | 4.8176 | 6.0562 | |
| | R2 | 0.6922 | 0.7202 | **0.7248** | 0.5311 | |
| | MAPE | 0.0585 | 0.0576 | **0.0570** | 0.0847 | |

*5.5 Conditional Average Treatment Effect of Traffic Crashes*

In this section, we applied EconML(Battocchi et al., 2019), a causal inference Python package that utilizes ML techniques, to estimate causal effects. We presented the results of CATE for each type of crash considering different spatial and temporal conditions, and we also discussed the validation of the results.

*5.5.1 Analysis and discussion of conditional average treatment effect results*

After we obtained $\hat{e}^{type}(x_{s,t})$ and $\hat{m}^{dur,dis}(t^{type}_{s,t}, x_{s,t})$, the final $\widehat{CATE}^{type}_{dur,dis}$ were estimated by OLS, and the results are outlined in Table 5, and we highlight the highest value for each distance, except those more than -1 mph. To conduct statistical hypothesis tests for validation, we apply the bootstrap method to calculate the p-value and confidence interval. The results show that the closer to the crash's time and location, the higher the confidence level, and when the effects are nearly zero, the p-values are high, meaning that there is no treatment effect of crashes.

In terms of the estimated CATE, it can be found that the degree of crash effect is REAR>WIPE>OBJ, which is consistent with previous findings(Garib et al., 1997; Chung, 2017). Regarding the spatial and temporal impact of crashes on highway traffic speed, our analysis reveals that crashes can cause an average speed reduction of around 10 to 15 miles per hour. Additionally, congestion propagates about 2-3 miles upstream and lasts over half an hour. We observed an apparent propagation and dissipation process of congestion, with different crash types exhibiting varying characteristics.

**Table 5. The estimated causal effects of crashes on speed (mph) for different duration and distance**

| Crash Type | dis / dur | 0 | -1 | -2 | -3 | -5 |
|---|---|---|---|---|---|---|
| REAR | 5 | -14.88***±2.78 | -9.31***±2.49 | -3.56***±1.95 | -0.73±1.51 | 0.36±0.75 |
|  | 10 | -15.88***±2.49 | -11.47***±3.06 | -3.66***±2.43 | -1.21**±1.66 | 0.48±0.74 |
|  | 15 | -15.63***±2.70 | -12.84***±3.42 | -6.51***±2.83 | -2.22±2.09 | 0.42±2.07 |
|  | 20 | -14.98***±2.74 | -12.38***±3.15 | -8.12***±3.14 | -2.38**±2.12 | 0.35±0.72 |
|  | 25 | -13.67***±2.63 | -10.55***±2.86 | -7.91***±3.24 | -1.75***±1.21 | 0.54±0.67 |
|  | 30 | -13.19***±2.87 | -10.16***±2.95 | -7.62***±2.97 | -2.32*±1.79 | 0.55±0.72 |
| OBJ | 5 | -9.09***±5.65 | -0.97±3.55 | -0.68±2.54 | -0.37±1.99 | -0.30±2.44 |
|  | 10 | -10.91*±6.83 | -1.81±4.21 | -0.79±2.86 | -0.39±1.71 | -0.31±2.31 |
|  | 15 | -9.91*±6.69 | -3.42±4.86 | -0.65±2.22 | -0.38±1.83 | -0.21±1.73 |
|  | 20 | -9.62±7.50 | -2.95±5.67 | -0.57±2.21 | -0.55±1.52 | -0.14±1.51 |
|  | 25 | -4.52*±3.22 | -2.02±5.99 | -0.68±1.81 | -0.35±1.77 | -0.14±1.59 |
|  | 30 | -3.61±4.03 | -1.27±4.14 | -0.71±4.18 | -0.45±1.98 | 0.00±1.97 |
| WIPE | 5 | -10.40***±3.63 | -2.06**±1.90 | -1.31±1.53 | -0.75±1.67 | 0.23±1.15 |
|  | 10 | -11.29***±4.51 | -3.25**±2.92 | -1.41*±1.32 | -0.55±1.47 | 0.46±1.30 |
|  | 15 | -12.08***±4.68 | -4.22***±3.89 | -1.72***±2.08 | -0.43±1.56 | 0.49±1.04 |
|  | 20 | -12.96***±4.75 | -4.67**±4.11 | -2.42*±1.92 | -0.79±1.69 | 0.23±1.11 |
|  | 25 | -11.46***±4.48 | -3.30*±4.20 | -2.33*±2.15 | -0.72±1.82 | 0.36±1.24 |
|  | 30 | -10.8***±4.90 | -2.80*±3.40 | -2.70±2.06 | -0.70±1.81 | 0.20±1.37 |

*Note: \*\*\*P<0.001; \*\*P<0.01; \*P<0.05; ±95%Confidence Interval*

Specifically, REAR crashes cause the most severe and lasting congestion compared to the other two crash types. The lowest speeds are observed at 0, -1, and -2 miles, occurring 10, 15, and 20 minutes after the occurrence time, respectively, demonstrating the propagation of a backward wave.

On the other hand, OBJ crashes contribute to the shortest congestion propagation distance since their CATE is nearly zero at -1 mile. OBJ crashes also exhibit the fastest recovery speed, with their CATE increasing from -10.91 to -3.61 mph within 30 minutes at 0 miles. This may be because OBJ mostly occurs at night, when fewer vehicles are on the road, and therefore, OBJ crashes do not affect many vehicles.

WIPE crashes result in slightly less severe congestion than REAR crashes but have the longest hysteresis of congestion. The maximum deceleration occurs 20 minutes after the occurrence of WIPE crashes, while it takes only 10 minutes for other crash types. This indicates that sideswipe crashes have obviously delayed impact and need to be addressed more timely and efficiently, such as by deploying clean-up crews and tow trucks.

The impact of time on the causal crash effect is explored in **Figure 10**. The peak hours (6:30-9:00 and 16:40-19:30) and night periods (23:00-4:00) are found to have different effects on crashes compared to off-peak hours (the rest time).According to the **Figure 10**, it can be found that the trends between REAR and OBJ in different periods are opposite, while the effect of WIPE remains stable. The underlying factors contributing to this phenomenon are that REAR and WIPE tend to occur during peak hour with higher volume. Additionally, due to the fact that average speed at night surpass those during peak hours, the impacts of REAR and WIPE are more pronounced during the nighttime hours. Conversely, OBJ is usually triggered by poor lighting conditions at night and tend to occur when traffic volume is relatively low. As a result, OBJ crashes occurring during peak hours exert a more substantial influence on average speeds. These findings provide insights into the heterogeneous effects of crashes on traffic speed, which can be valuable for policymakers in developing effective traffic management strategies.

*5.5.2 Validation of estimated causal effect*

This study validates the proposed method from two aspects to demonstrate its rationality. Firstly, **Figure 11** illustrates the change process of traffic speeds affected by crashes through several individual examples, where the occurrence times of crashes are set as 0 on the x-axis. The orange lines represent the actual speeds before and after crashes, and the green lines represent the average speed without crash treatment, determined from the traffic condition *one week* after the crashes. The green area denotes the individual treatment effect since the effect is inferred for a single crash. The figure shows that the counterfactual speeds demonstrate reasonable trends and quantities compared to average speeds, proving that the proposed method provides an acceptable estimation of causal effects.

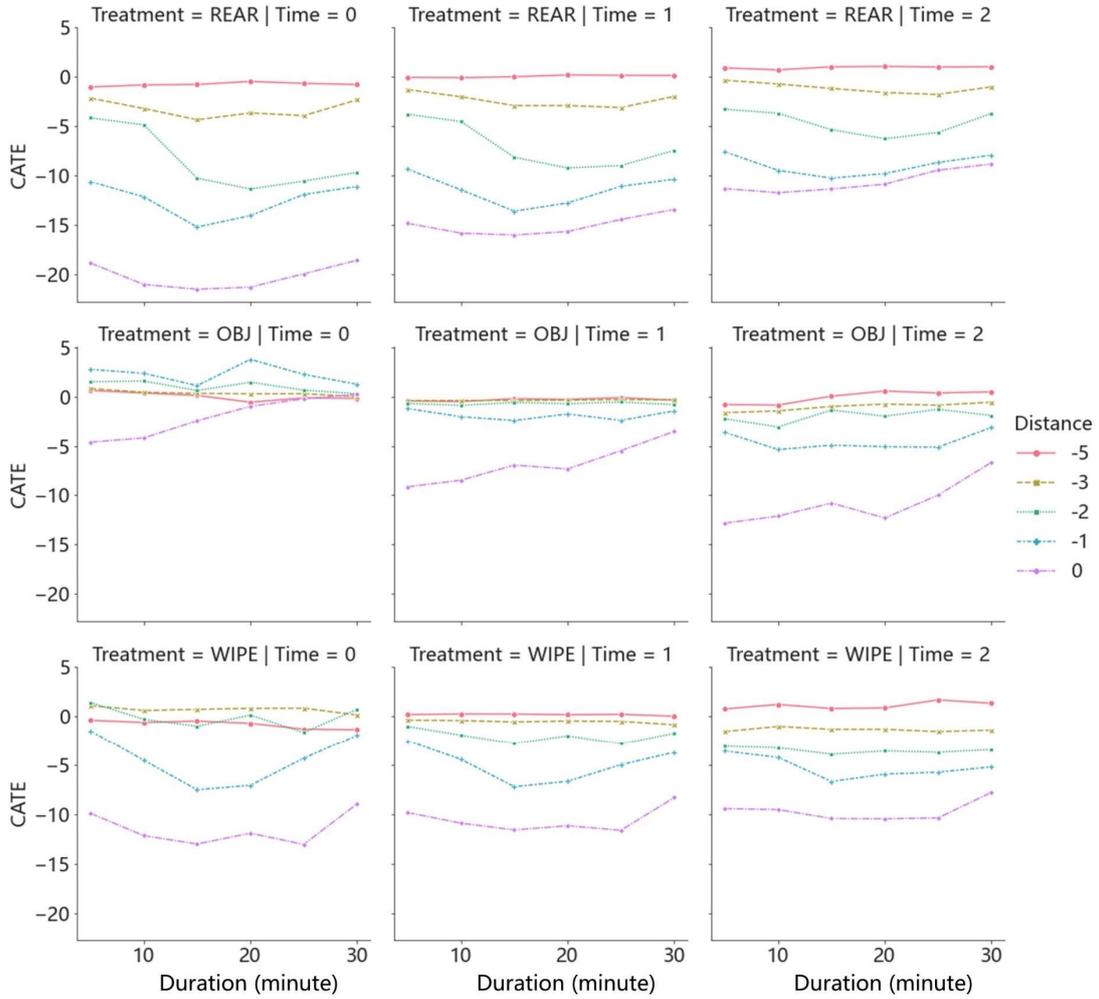

**Figure 10. Comparison of CATE in different time periods (0: night, 1: off-peak, 1: peak hour)**

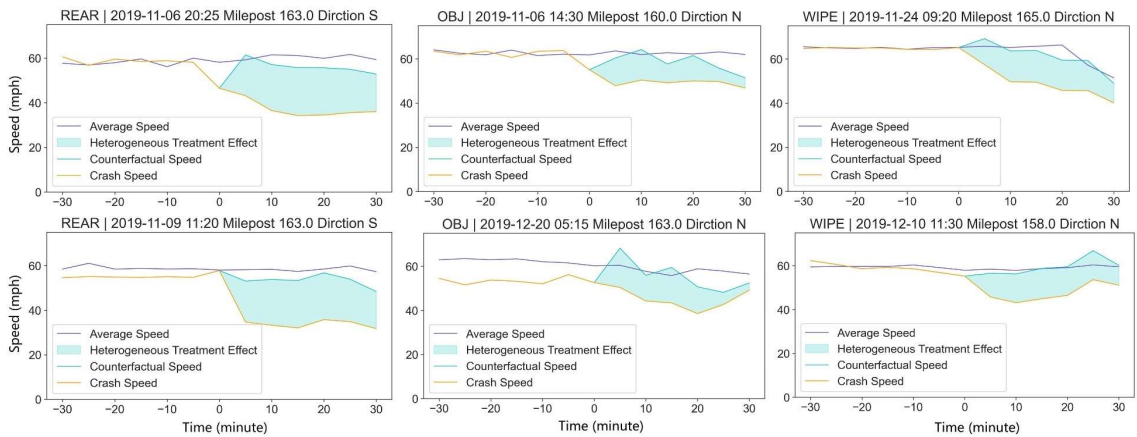

**Figure 11. Examples of individual treatment effects compared by matched average speed. The green areas between green and orange lines refer to estimated effects.**

Secondly, we evaluate the overall estimate performance by calculating error metrics based on the matched "counterfactual outcomes" generated by matching algorithm. Furthermore, we used GRF and DML as comparisons, as shown in Table 6. The main difference between GRF and DRL lies in their use of decision trees. GRF is an extension of the traditional random forest algorithm that uses a double-robust estimation procedure to adjust for confounding variables. As for DML, it also estimates the causal effects by two stages process but predicts the outcome from just the controls rather than both treatment and controls in the first stage. Because DML does not use double-robust estimation, more bias may be introduced into the final regression. To calculate the error metrics, we substituted the estimated value $\hat{y}_i$ and actual value $y_i$ in Equation (19)-(22) with $ice_i$ and $mce_i$ (Equation (14)), respectively. We found that the DRL model outperforms the other methods in terms of accuracy. This further validates the rationality and effectiveness of our proposed method.

**Table 6. Performance comparison based on matched data.**

| Methods | MAE | MSE | RMSE | MAPE |
| --- | --- | --- | --- | --- |
| GRF | 12.60 | 345.18 | 17.76 | 0.22 |
| DML | 12.81 | 322.54 | 16.88 | 0.24 |
| DRL | **10.73** | **244.59** | **14.76** | **0.20** |

*5.5.3 Validation of variables selection*

We validated the effectiveness of variable selection by sensitive analyses from two perspectives. Firstly, we compared the estimated results and error performance before and after the selection process by DRL. **Figure 12** illustrates that without variable selection, the CATE is nearly zero, which is inconsistent with real-world phenomena and indicates the failure of causal inference. Table 7 shows the error performance and validates the improvement due to variable selection. Additionally, when estimating the errors for different times and locations, **Figure 13** indicates that the effectiveness of variable selection reduces as time and distance increase. This means that the proposed method has no advantages in terms of spatial and temporal causal inference.

**Table 7. Average performance of DRL before and after variables selection**

| Type | Method | MAE | MSE | RMSE | MAPE |
| --- | --- | --- | --- | --- | --- |
| REAR | Before selection | 10.74 | 259.12 | 15.01 | 0.21 |
| | After selection | **9.17** | **158.94** | **12.37** | **0.17** |
| OBJ | Before selection | 12.6 | 375.93 | 17.7 | 0.23 |
| | After selection | **11.01** | **367.72** | **16.3** | **0.21** |
| WIPE | Before selection | 11.40 | 293.48 | 15.99 | 0.23 |
| | After selection | **10.03** | **203.2** | **13.66** | **0.18** |

Secondly, we validated the estimated performance based on different threshold values, ranging from 0 to 0.40, with a 0.05 step. The average error performance is presented in Table 8. Based on the results, it is evident that the optimal performance is achieved when the threshold is set to 0.15.

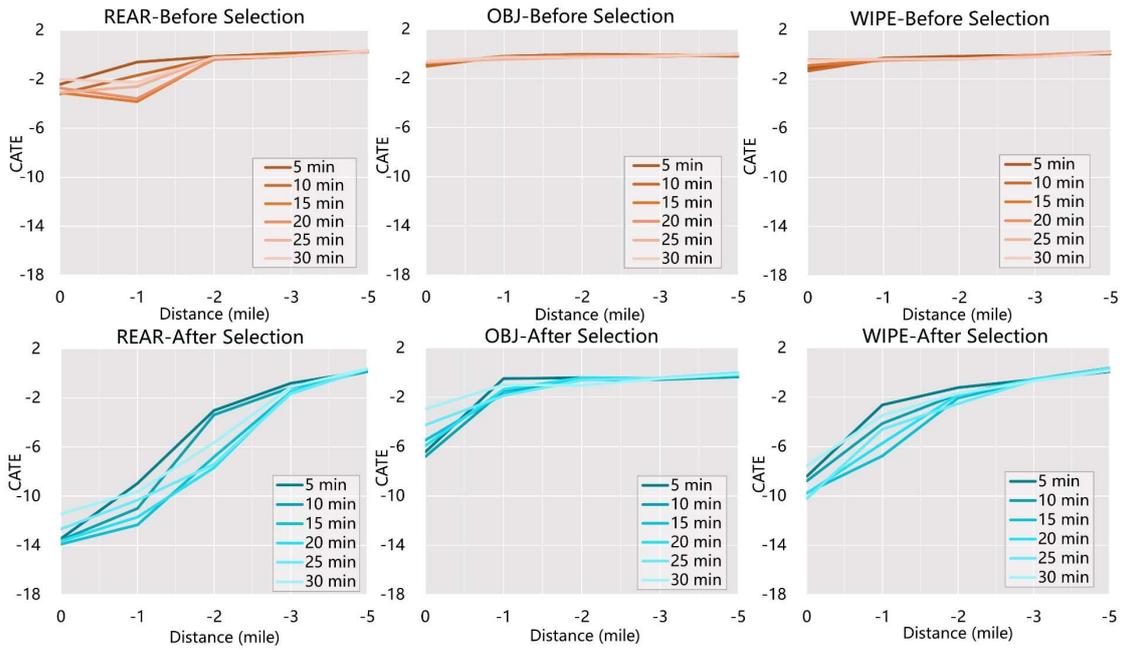

**Figure 12. Estimated CATE before and after variables selection. Each line refers to the CATE of crashes 5-30 min after.**

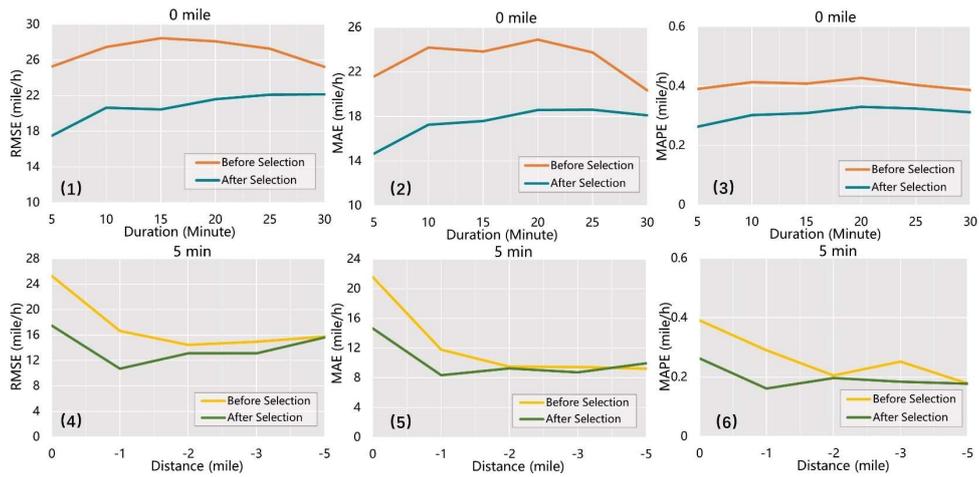

**Figure 13. Estimated error before and after variables selection. (1-3) are the errors of crash sites changing with time; (4-6) are the errors 5 minutes after crashes changing with distance.**

**Table 8. The average error performance for different threshold of CSVI**

| Threshold | MAE | MSE | RMSE | MAPE |
|---|---|---|---|---|
| 0.00 | 7.82 | 129.26 | 10.49 | 0.1380 |
| 0.05 | 7.81 | 129.27 | 10.49 | 0.1379 |
| 0.10 | 7.79 | 128.57 | 10.46 | 0.1376 |
| **0.15** | **7.77** | **127.98** | **10.44** | **0.1372** |
| 0.20 | 7.82 | 129.38 | 10.50 | 0.1381 |
| 0.25 | 7.80 | 128.96 | 10.49 | 0.1378 |
| 0.30 | 7.85 | 130.38 | 10.58 | 0.1383 |
| 0.35 | 8.33 | 135.32 | 11.01 | 0.1462 |
| 0.40 | 8.56 | 133.52 | 10.92 | 0.1496 |

## 6. Conclusions

This paper presents a novel causal machine learning framework for estimating heterogeneous causal effect of traffic crashes on speed reduction. The framework utilizes Doubly Robust Learning models, which combine machine learning and doubly robust inference methods. Additionally, the framework incorporates causal inference theory and notations to guide variable selection and causal structure analysis. In order to showcase the effectiveness and precision of the proposed framework, real-world data encompassing incident records, traffic flow information, and road geometry details were utilized from the Interstate 5 freeway located in Washington.

The results of the estimated CATE by proposed method reveal heterogeneous effects of crashes on speed. Overall, REAR exhibits the most significant impact, followed by WIPE, and finally OBJ. The estimated effects also reveal the propagation and dissipation mechanism of traffic congestion: REAR has a greater impact on traffic further upstream, whereas the influence range of OBJ is the shortest. The delayed effects of WIPE are particularly pronounced. Furthermore, REAR exhibits more adverse effects at night, whereas OBJ demonstrates more adverse effects during peak hours, which can be attributed to the road conditions. The proposed method also enables the inference of individual treatment effects of crashes on speed, supported by comprehensive experiments that validate its effectiveness and accuracy. These findings can significantly assist authorities in formulating timely emergency policies for highway safety.

This study offers a new perspective for analyzing the causality between highway crashes and traffic speed. However, there are still avenues for future research, including analyzing the causality of other factors that are hard to observe in normal traffic flow, such as drinking, gender, and age. Additionally, the proposed model cannot deal with the spatial and temporal causal relationship since traffic data are time series and have spatial dependencies. Therefore, modifying the causal inference methods to improve the ability for spatiotemporal data would be worthwhile.

**CRediT authorship contribution statement**

**Shuang Li:** Methodology, Software, Data curation, Visualization, Writing - original draft. **Ziyuan Pu:** Conceptualization, Methodology, Supervision, Resources, Investigation, Writing - original draft. **Zhiyong Cui:** Conceptualization, Methodology, Supervision, Investigation, Writing - review &

editing. **Seunghyeon Lee:** Conceptualization, Methodology, Writing - review & editing. **Xiucheng Guo:** Conceptualization, Methodology, Writing - review & editing. **Dong Ngoduy:** Conceptualization, Methodology, Writing - review & editing.

**Declaration of Competing Interest**

The authors declare that they have no known competing financial interests or personal relationships that could have appeared to influence the work reported in this paper.

**Acknowledgments**

This work was partially supported by the National Natural Science Foundation of China (project number: 52202378), the Postgraduate Research&Practice Innovation Program of Jiangsu Province (project number: KYCX22_0271), the Ministry of Transport of PRC Key Laboratory of Transport Industry of Comprehensive Transportation Theory (Nanjing Modern Multimodal Transportation Laboratory) (project number: MTF2023002).